\documentclass[letterpaper, 10 pt, conference]{ieeeconf}  

\IEEEoverridecommandlockouts                              

\usepackage{graphicx}
\usepackage{amsmath}
\usepackage{amssymb}
\usepackage{mathtools}
\usepackage{array}
\usepackage{booktabs}
\usepackage{multirow}
\usepackage{tabularx}
\usepackage{url}
\usepackage[capitalise]{cleveref}

\crefname{equation}{}{}
\Crefname{equation}{}{}

\DeclareMathOperator*{\argmin}{arg\,min}

\newif\ifanonym
\anonymfalse   

\title{\LARGE \bf
CN-CBF: Composite Neural Control Barrier Function for Robot Navigation in Dynamic Environments
}

\ifanonym
    \author{}
\else
    \author{Bojan Derajić$^{1}$, Sebastian Bernhard$^{2}$ and Wolfgang Hönig$^{1, 3}$
    \thanks{$^{1}$Technical University of Berlin, Germany}%
    \thanks{$^{2}$AUMOVIO, Germany}%
    \thanks{$^{3}$Robotics Institute Germany (RIG), Germany}%
    \thanks{Contact e-mail: {\tt\small bojan.derajic@campus.tu-berlin.de}}%
    \thanks{This research is supported by the German Federal Ministry for Economic Affairs and Energy within the project "NXT GEN AI METHODS – Generative Methoden für Perzeption, Prädiktion und Planung".}%
    }
\fi

\begin{document}

\maketitle
\thispagestyle{empty}
\pagestyle{empty}

\begin{abstract}

Safe navigation of autonomous robots remains one of the core challenges in the field, especially in dynamic and uncertain environments. One prevalent approach is safety filtering based on control barrier functions (CBFs), which are easy to deploy but difficult to design. Motivated by the shortcomings of existing learning- and model-based methods, we propose a simple yet effective neural CBF design method for safe robot navigation in dynamic environments. We employ the idea of a composite CBF, where multiple neural CBFs are combined into a single CBF. Individual CBFs are trained using data generated offline via the Hamilton-Jacobi reachability framework to approximate the optimal safe set for single moving obstacles. Additionally, we use a residual neural architecture, ensuring that the estimated safe set does not intersect with the corresponding failure set. The method is extensively evaluated in simulation experiments for a ground robot and a quadrotor, comparing it against several baseline methods. The proposed method improves success rates by up to 18\% over the strongest baseline, while maintaining comparable or lower path lengths and motion times. The method is also demonstrated in hardware experiments for both types of robots.

\end{abstract}

\section{INTRODUCTION}
\label{sec:introduction}

Motion planning and control of mobile robots in dynamic and unknown environments has been an active research field for decades and many advanced methods have been published. Safety filtering is one of the most prominent and well-established approaches to safety-critical scenarios. This approach functions by monitoring the nominal control and modifying it when necessary to preserve safe operation of the system. The nominal control is usually modified by a switching mechanism or through an optimization procedure \cite{hsu_safety_2024}. In this paper, we follow the second approach, where we specifically utilize control barrier functions (CBFs).

CBFs are scalar functions defined on the system's state space characterizing safe control invariant sets based on Nagumo's Theorem on set invariance \cite{nagumo_uber_1942, ames_control_2019}. In practice, the main advantage of using CBFs is highlighted for control-affine systems, where the safety filtering is reduced to solving a quadratic program (QP) and can be executed at a high rate \cite{ames_control_2014}. Nevertheless, designing a proper CBF and finding computationally tractable, data-efficient methods remain ongoing challenges \cite{hsu_safety_2024}. Also, an additional layer of complexity is present for systems such as mobile robots that operate in dynamic and uncertain environments based on local perception. In such cases, the CBF must be generated in real time from the available information, rendering the existing methods for offline CBF design ineffective \cite{lin_one_2026}.

In this paper, we propose a new approach to designing a CBF for mobile robots operating in dynamic environments. The goal is to safely navigate the environment without colliding with any obstacle, corresponding to the robot's state entering a \textit{failure set}. We first design a neural CBF for a single robot-obstacle pair and then apply an aggregation function that outputs a single composite CBF for an arbitrary number of individual CBFs. To recover the optimal safe sets for individual obstacles, as a CBF we use a neural approximation of the value function obtained offline via Hamilton-Jacobi (HJ) reachability analysis. Importantly, we apply HJ reachability to the relative dynamics of the robot and the obstacle rather than the robot dynamics directly, which results in the failure set being stationary instead of time-varying. Moreover, by using a residual neural architecture, we ensure that the estimated safe set of individual CBFs, as well as the composite CBF, never intersects the corresponding set of failure states. In summary, our contributions are:
\begin{itemize}
    \item A neural CBF design method for dynamic environments that simultaneously recovers near-optimal safe sets for individual obstacles, guarantees the estimated safe set never intersects the failure set, and scales to an arbitrary number of obstacles — none of which is achieved jointly by existing approaches. \\
    \item Extensive evaluation of the proposed method against several relevant baselines in simulation experiments and hardware demonstration for two types of robots - a ground robot and a quadrotor.  
\end{itemize}

\section{RELATED WORK}
\label{sec:related_work}

When it comes to safety-critical mobile robot deployment, two types of methods are prevalent: receding horizon approaches and safety filtering \cite{hwang_safe_2024}. The first category comprises different MPC-based local planners (e.g., \cite{brito_model_2019, samavi_sicnav_2025, derajic_residual_2025}), while the second is dominated by CBF-based methods \cite{hsu_safety_2024}, which we discuss further.

There is a significant amount of literature on CBF-based methods as they have been proven to be a powerful, yet elegant way to enforce safety \cite{ames_control_2019}. Still, designing a proper CBF for a general nonlinear system with state and control constraints remains the major challenge \cite{hsu_safety_2024}. The most relevant studies leverage learning techniques to design CBFs more flexibly \cite{dawson_safe_2023}. For example, learning CBFs from expert trajectories is proposed in \cite{robey_learning_2020}, while the neural CBFs  for systems with input limits are introduced in \cite{liu_safe_2022}. Also, designing CBFs in real time via Gaussian splatting is used in \cite{chen_control_2025}. However, these approaches often fail to recover the optimal safe set, which can be obtained via HJ reachability analysis \cite{so_how_2024}.  

The connection between CBFs and HJ reachability is established in \cite{choi_robust_2021}, where the novel control-barrier value function (CBVF) is introduced as a method for the CBF design, but demonstrated only for known environments. Overcoming this disadvantage, the CBFs in \cite{derajic_orn-cbf_2026} and \cite{lin_one_2026} are observation-conditioned neural methods grounded in HJ reachability theory and approximately recover the maximal safe set. The first uses a residual architecture to ensure that the predicted safe set never intersects occupied space, while the second includes uncertainty estimation. However, these methods do not consider dynamic obstacles explicitly. A neural approximation of a joint HJ value function for moving obstacles is used in \cite{derajic_residual_2025}, but is limited to a low number of obstacles due to combinatorial complexity. Instead, we apply HJ reachability to dynamic environments by aggregating multiple HJ functions into a single function, reducing the total time required for data generation and model training from dozens of hours to minutes.

The idea of composite CBFs has been explored in several previous works. The authors in \cite{harms_safe_2025} propose a composite CBF tailored for safe quadrotor navigation only in static environments. Time-varying safety constraints are considered in \cite{safari_time-varying_2024}, but the safe control is provided in a closed form without imposing actuation limits. The method proposed in \cite{yu_sequential_2023} learns interaction dynamics between the robot and an obstacle from data and uses a non-smooth aggregation function, resulting in a complicated training process, a suboptimal safe set, and a non-smooth composite CBF. On the other hand, we introduce a smooth, learning-based composite CBF approach that handles dynamic obstacles while respecting the control limits of the robot's model.       


\section{PRELIMINARIES}
\label{sec:preliminaries}

In this section, we present a brief overview of the two frameworks serving as the basis for our approach --- CBFs and HJ reachability.  We assume a continuous-time setting, but for readability, we omit the time variable $t$ where it is not crucial.

\subsection{Control Barrier Functions}
\label{subsec:cbf}
We consider control-affine system dynamics
\begin{equation} \label{eq:system_dynamics}
    \dot{\mathrm{x}} = f(\mathrm{x}) + g(\mathrm{x})\mathrm{u}
\end{equation}
where $\mathrm{x} \in \mathcal{X} \subset \mathbb{R}^{n_x}$ and $\mathrm{u} \in \mathcal{U} \subset \mathbb{R}^{n_u}$. A continuously differentiable function $h : \mathcal{X} \times [0, \infty) \rightarrow \mathbb{R}$ is a valid control barrier function if there exists an extended class $\kappa$ function $\alpha : \mathbb{R} \rightarrow \mathbb{R}$ such that
\begin{equation} \label{eq:cbf_constr}
    \sup_{\mathrm{u} \in \mathcal{U}} \nabla_{\mathrm{x}}h(\mathrm{x}, t)^\top \dot{\mathrm{x}} + \frac{\partial h(\mathrm{x}, t)}{\partial t} \geq -\alpha\left(h(\mathrm{x}, t)\right).
\end{equation}
If we define the set $\mathcal{S}$ as
\begin{equation}
    \mathcal{S}(t) = \{ \mathrm{x} \in \mathcal{X} : h(\mathrm{x}, t) \geq 0 \},
\end{equation}
then $\mathcal{S}(t)$ is forward invariant and we call it \textit{safe set} \cite{lindemann_control_2019}.

When the admissible control set $\mathcal{U}$ can be expressed through linear constraints, the CBF condition \cref{eq:cbf_constr} can be leveraged to construct a QP-based safety filter \cite{ames_control_2014}. Given the current state $\mathrm{x}$ and a nominal input $\mathrm{u}_{\text{nom}}$, provided that the QP is feasible, the CBF-QP returns the admissible control $\mathrm{u}^*$ closest to $\mathrm{u}_{\text{nom}}$ that satisfies the CBF constraint:
\begin{equation} \label{eq:cbf_qp}
\begin{aligned}    
    &\mathrm{u}^* = \argmin_{\mathrm{u} \in \mathcal{U}} \,\, \frac{1}{2} \| \mathrm{u}_{\text{nom}} - \mathrm{u} \|^2 \\
    & s.t. \quad L_{f}h(\mathrm{x}, t) + L_{g}h(\mathrm{x}, t) \mathrm{u} + \frac{\partial h(\mathrm{x}, t)}{\partial t} + \alpha \left( h(\mathrm{x}, t) \right)  \geq 0, 
\end{aligned}
\end{equation}
where $L_{f}h(\mathrm{x}, t)$ and $L_{g}h(\mathrm{x}, t)$ are the Lie derivatives with respect to $f(\mathrm{x})$ and $g(\mathrm{x})$, respectively.

\subsection{Hamilton-Jacobi Reachability Analysis}
\label{subsec:hjr}

HJ reachability is a framework for verification and analysis of dynamical systems based on the optimal control theory \cite{bansal_hamilton-jacobi_2017}. In general, here we consider a general system dynamics
\begin{equation}
    \dot{\mathrm{x}} = f(\mathrm{x}, \mathrm{u}, \mathrm{d}),
\end{equation}
where $\mathrm{u} \in \mathcal{U} \subset \mathbb{R}^{n_u}$ and $\mathrm{d} \in \mathcal{D} \subset \mathbb{R}^{n_d}$ are two competing inputs. If the system starts at state $\mathrm{x}(t)$, then $\xi^{\mathrm{u},\mathrm{d}}_{\mathrm{x}, t}(\tau)$ is the system's state at time $\tau$ after applying $\mathrm{u}(\cdot)$ and $\mathrm{d}(\cdot)$ over time horizon $[t, \tau]$. The set of states that should be ultimately avoided is called a failure set $\mathcal{F}$. Also, we define a backward reachable tube (BRT) $\mathcal{B}(t)$, which is a set of initial states for which exists $\mathrm{d}(\cdot)$ such that the system will reach $\mathcal{F}$ within the time horizon $[t, T]$ for any control $\mathrm{u}(\cdot)$, i.e. 
\begin{equation}
    \mathcal{B}(t) = \{ \mathrm{x}: \exists \mathrm{d}(\cdot), \forall \mathrm{u}(\cdot), \, \exists \tau \in [t, T], \, \xi^{\mathrm{u},\mathrm{d}}_{\mathrm{x}, t}(\tau)  \in \mathcal{F} \}.
\end{equation}

To compute the BRT for a given $\mathcal{F}$, we define it as the zero-sublevel set of a failure function $\ell(\mathrm{x})$, i.e. ${\mathcal{F} = \{\mathrm{x} : \ell(\mathrm{x}) \leq 0 \}}$. The computation of the BRT can be formulated as a min-max optimization problem where the objective is the minimal distance to $\mathcal{F}$ over the time horizon:
\begin{equation}
    J(\mathrm{x}, \mathrm{u}(\cdot), \mathrm{d}(\cdot), t) = \min_{\tau \in [t, T]} \ell(\xi^{\mathrm{u},\mathrm{d}}_{\mathrm{x}, t}(\tau)).
\end{equation}
The control input $\mathrm{u}$ keeps the system trajectory away from the failure region $\mathcal{F}$, while  the adversarial input $\mathrm{d}$ pushes it toward $\mathcal{F}$. This can be described by the value function 
\begin{equation} \label{eq:hj_value_func}
    V(\mathrm{x}, t) = \inf_{\mathrm{d}(\cdot) \in \mathcal{D}}  \sup_{\mathrm{u}(\cdot) \in \mathcal{U}} \{ J(\mathrm{x}, \mathrm{u}(\cdot), \mathrm{d}(\cdot), t) \}.
\end{equation}
This value function can be computed using the dynamic programming principle to solve the following Hamilton-Jacobi-Isaacs Variational Inequality \cite{bansal_hamilton-jacobi_2017}:
\begin{equation} \label{eq:hjb_vi}
    \begin{aligned}
    \min \left\{ \frac{\partial}{\partial t}V(\mathrm{x}, t) + H(\mathrm{x}, t), \, \ell(\mathrm{x}) - V(\mathrm{x}, t) \right\} &= 0, \\
    V(\mathrm{x}, T) &= \ell(\mathrm{x}).
    \end{aligned}
\end{equation}
In the formulation above, Hamiltonian $H(\mathrm{x}, t)$ is defined as
\begin{equation}
    H(\mathrm{x}, t) = \max_{\mathrm{u} \in \mathcal{U}} \min_{\mathrm{d} \in \mathcal{D}} \nabla V(\mathrm{x}, t)^\top f(\mathrm{x}, \mathrm{u}, \mathrm{d}).
\end{equation}
Once the value function \cref{eq:hj_value_func} is obtained, the corresponding BRT is zero-sublevel set of $V(\mathrm{x}, t)$, i.e.
\begin{equation}
    \mathcal{B}(t) = \left\{ \mathrm{x}: V(\mathrm{x}, t) \leq 0 \right\},
\end{equation}
and the maximal safe set is its complement: $\mathcal{S}(t) = \mathcal{B}(t)^c$.

Assuming terminal time $T=0$ and static $\ell(\mathrm{x})$, we propagate the value function backward in time until it converges to its steady-state value $V(x) = \lim_{t\rightarrow-\infty}V(x, t)$ and then use $V(x)$ as the target CBF during training.

\section{METHODOLOGY}
\label{sec:methodology}

In this section, we present a detailed description of the proposed CN-CBF method. We first clarify the assumed structure of the problem, and then proceed with further explanation. Our focus is on dynamic environments where safety constraints evolve over time. This condition is explicitly taken into account in \cref{eq:cbf_constr} as a partial time derivative of the CBF, representing a change of the safe set due to the motion of obstacles. We first consider a scenario with a single moving obstacle, and later extend the method for an arbitrary number of obstacles.

Obtaining a CBF with the optimal safe set for the single obstacle case is possible via HJ reachability framework \cite{so_how_2024, hsu_safety_2024}. A direct approach would be to define a time-dependent failure function $\ell(\mathrm{x}, t)$ based on the motion of the obstacle, which is generally impractical. Instead, we consider a moving obstacle as a dynamical system with a state vector $\mathrm{o} \in \mathcal{O} \subset \mathbb{R}^{n_o}$ whose dynamics evolve according to
\begin{equation} \label{eq:obstacle_dynamics}
    \dot{\mathrm{o}} = \nu(\mathrm{o}, \mathrm{d}),
\end{equation}
where $\mathrm{d} \in \mathcal{D} \subset \mathbb{R}^{n_d}$ is the input to the system. Next, since only the relative motion of the robot and the obstacle matters, we define a transformation $\rho : \mathbb{R}^{n_x} \times \mathbb{R}^{n_o} \rightarrow \mathbb{R}^{n_z}$ which outputs a relative state vector $\mathrm{z} \in \mathcal{Z} \subset \mathbb{R}^{n_z}$, i.e.
\begin{equation} \label{eq:rho}
    \mathrm{z} = \rho(\mathrm{x}, \mathrm{o}).
\end{equation}
The relative dynamics is then obtained as
\begin{equation} \label{eq:relative_dynamics}
\begin{aligned}
    \dot{\mathrm{z}} &= \frac{\partial \mathrm{z}}{\partial\mathrm{x}} \dot{\mathrm{x}} + \frac{\partial \mathrm{z}}{\partial\mathrm{o}} \dot{\mathrm{o}} \\
    &= \zeta(\mathrm{z}, \mathrm{u}, \mathrm{d}).
\end{aligned}
\end{equation}
The transformation $\rho$, i.e. relative state $\mathrm{z}$, is a design choice and depends on the concrete robot and (assumed) obstacle dynamics. Typically, the relative state contains relative pose and velocity. In \cref{subsec:experiments} we provide concrete examples for a ground robot and a quadrotor navigating in a crowd.

The advantage of using a relative state representation in such a setting is twofold. First, the failure function $\ell(\mathrm{z})$ defined for the relative state is stationary because a fixed set of relative states is considered a failure regardless of the absolute motion of the robot and the obstacle. Second, the relative dynamics \cref{eq:relative_dynamics} has two competing inputs --- the robot's input $\mathrm{u}$ driving the relative state from the failure region, and the obstacle's input $\mathrm{d}$ driving it toward the failure region. This makes \cref{eq:relative_dynamics} exactly the type of system considered in the HJ reachability framework, allowing the application of existing tools.  

Motivated by the discussion above, we design a CBF for the relative dynamics rather than the original robot dynamics \cref{eq:system_dynamics}. Our aim is to use the value function obtained by HJ reachability as a relative CBF with the maximal safe set. However, numerical tools for HJ reachability provide only the numerical values of the value function on the predefined state-space grid. This means that the deployed CBF would require considerable memory to store these values and an additional interpolation logic to obtain the value and gradient for a given state. Instead, we use a neural approximation of the value function, allowing for lower memory requirements and efficient value and gradient queries via automatic differentiation.

Besides, we recall that the HJ value function defined in \cref{eq:hj_value_func} is less than or equal to the failure function, so for relative dynamics we have $V(\mathrm{z}, t) \leq \ell(\mathrm{z}), \quad \forall \mathrm{z} \in \mathcal{Z}, \; \forall t$ \cite{derajic_residual_2025, derajic_orn-cbf_2026}. Therefore, assuming that the value function converged in time, we can write $V(\mathrm{z}) = \ell(\mathrm{z}) - r(\mathrm{z})$ where ${r(\mathrm{z}) \geq 0, \quad \forall \mathrm{z} \in \mathcal{Z}}$ is the residual function. Since $\ell(\mathrm{z})$ is usually defined as a signed distance function that can be obtained efficiently in real time, we can approximate only the residual component. By defining the relative CBF as $\bar{h}(\mathrm{z}) \coloneqq V(\mathrm{z})$ and approximating the residual function with a neural model with parameters $\Theta$, we have   
\begin{equation} \label{eq:h_bar}
    \bar{h}_{\Theta}(\mathrm{z}) = \ell(\mathrm{z}) - r_{\Theta}(\mathrm{z}).
\end{equation}
Moreover, if the neural model has a non-negative activation function at the output, i.e. $r_{\Theta}(\mathrm{z}) \geq 0, \quad \forall \mathrm{z} \in \mathcal{Z}$, it is guaranteed that the learned safe set of $\bar{h}_{\Theta}(\mathrm{z})$ never intersects the failure set defined by $\ell(\mathrm{z})$ \cite{derajic_orn-cbf_2026}.

Nevertheless, to make the learned CBF usable, it is necessary to establish a relation between $\bar{h}_{\Theta}(\mathrm{z})$ and the initial CBF constraint defined for robot dynamics \cref{eq:cbf_constr}. If we consider relative dynamics \cref{eq:relative_dynamics}, we have
\begin{subequations}\label{eq:h_bar_to_h}
\begin{align}
    h(\mathrm{x}, t) &= \bar{h}_{\Theta}(\mathrm{z}) \label{eq:h_bar_to_h:a}\\
    \nabla_{\mathrm{x}}h(\mathrm{x}, t)^\top &= \nabla_{\mathrm{z}}\bar{h}_{\Theta}(\mathrm{z})^\top \frac{\partial\mathrm{z}}{\partial\mathrm{x}} \label{eq:h_bar_to_h:b}\\
    \frac{\partial h(\mathrm{x}, t)}{\partial t} &= \nabla_{\mathrm{z}}\bar{h}_{\Theta}(\mathrm{z})^\top \frac{\partial\mathrm{z}}{\partial\mathrm{o}}
    \dot{\mathrm{o}} \label{eq:h_bar_to_h:c}
\end{align}
\end{subequations}
which completely defines the constraint in \cref{eq:cbf_constr} and enables safe control. From the equations above, it is evident that the partial time derivative of the CBF is nonzero solely due to obstacle dynamics, which aligns with our expectations.

The next step is to extend the proposed method for multiple obstacles. To do so, we use the idea of the composite CBF where multiple CBFs are designed (typically one per obstacle) and combined into a single CBF using some aggregation function \cite{molnar_composing_2023, harms_safe_2025}. Therefore, if there are $M$ obstacles with states $\{ \mathrm{o}_i\}_{i=1}^M$, for the given robot state $\mathrm{x}$ we first obtain the set of relative states $\{ \mathrm{z}_i \}_{i=1}^M$ and compute the corresponding relative CBFs $\{ \bar{h}_{\Theta}(\mathrm{z}_i) \}_{i=1}^M$. Then, we simply apply the aggregation function $\eta : \mathbb{R}^M \rightarrow \mathbb{R}$ to obtain a single CBF value, i.e.
\begin{equation} \label{eq:cn_cbf}
    h_{\Theta}(\mathrm{z}_1, \ldots, \mathrm{z}_M) = \eta\left(\bar{h}_{\Theta}(\mathrm{z}_1), \ldots, \bar{h}_{\Theta}(\mathrm{z}_M)\right).
\end{equation}
In this paper, as the aggregation function, we use a smooth under-approximation of the min function \cite{lindemann_control_2019}:
\begin{equation} \label{eq:eta}
    \eta\left(\bar{h}_{\Theta}(\mathrm{z}_1), \ldots, \bar{h}_{\Theta}(\mathrm{z}_M)\right) = -\frac{1}{\beta} \ln \sum_{i=1}^M e^{-\beta\bar{h}_{\Theta}(\mathrm{z}_i)},
\end{equation}
where $\beta \in \mathbb{R}_+$ is the smoothing parameter. For a fixed number of obstacles $M$, this choice of $\eta$ ensures that the composite BRT is a superset of the union of individual BRTs and yields a differentiable composite CBF \cite{molnar_composing_2023}.

Similarly to the single-obstacle case, we can establish a relation between the composite CBF and the CBF constraint in \cref{eq:cbf_constr} as:
\begin{subequations}\label{eq:h_cn_to_h}
\begin{align}
    h(\mathrm{x}, t) &= h_{\Theta}(\{\mathrm{z}\}_{i=1}^M) \label{eq:h_cn_to_h:a}\\
    \nabla_{\mathrm{x}}h(\mathrm{x}, t)^\top &= \sum_{i=1}^M\nabla_{\mathrm{z}_i}h_{\Theta}(\{\mathrm{z}\}_{i=1}^M)^\top \frac{\partial \mathrm{z}_i}{\partial\mathrm{x}} \label{eq:h_cn_to_h:b}\\
    \frac{\partial h(\mathrm{x}, t)}{\partial t} &= \sum_{i=1}^M\nabla_{\mathrm{z}_i}h_{\Theta}(\{\mathrm{z}\}_{i=1}^M)^\top \frac{\partial \mathrm{z}_i}{\partial\mathrm{o}_i} \dot{\mathrm{o}}_i \label{eq:h_cn_to_h:c}
\end{align}
\end{subequations}

\begin{figure}
    \centering
    \includegraphics[width=0.8\linewidth]{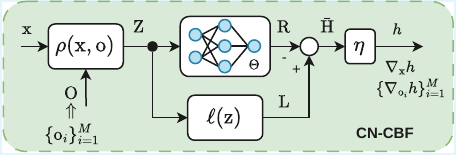}
    \caption{Structure of the proposed CN-CBF method.}
    \label{fig:cn_cbf}
\end{figure}

An important practical note is that if all the functions are implemented using a framework that supports automatic differentiation (e.g., PyTorch), the CBF value and its gradients can be obtained in a single forward and backward pass. This is important since the derivation of the analytical expressions for gradients might be overly complex. First, the set of obstacle states $\{ \mathrm{o}_i \}_{i=1}^M$ is converted into a batch of vectors represented as a tensor $\mathrm{O} \in \mathbb{R}^{M\times n_o}$. For the current robot state $\mathrm{x}$ and based on $\mathrm{O}$, we form a tensor of relative states $\mathrm{Z} \in \mathbb{R}^{M\times n_z}$ which is propagated through the neural model to obtain tensor of residuals values $\mathrm{R} \in \mathbb{R}^{M}$, while in parallel a tensor of failure function values $\mathrm{L} \in \mathbb{R}^{M}$ is computed. Then, by subtracting $\mathrm{R}$ from $\mathrm{L}$ we get $\bar{\mathrm{H}}$ and finally by applying $\eta$ the composite CBF value is computed. We call this method composite neural CBF (CN-CBF) and the described procedure is visualized in \cref{fig:cn_cbf}.

It should be noted that the number of obstacles $M$ is not predefined and can change over time, making the approach particularly flexible and scalable. However, whenever the number $M$ changes, there is a discontinuity in the composite CBF that might affect the behavior of the safety filter. From the aspect of safety, we are interested in the case when $M$ increases, as it enlarges the unsafe region and states that were safe suddenly become unsafe. Therefore, to ensure continual safety, it is required that the newly observed obstacles do not render the current robot's state unsafe instantly \cite{derajic_orn-cbf_2026}. In practice, this can be ensured to some extent by using a sensor with a sufficiently large perception field.

\begin{figure}
    \centering
    \includegraphics[width=0.85\linewidth]{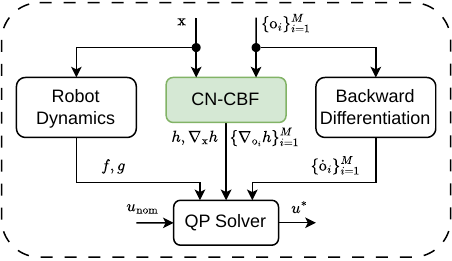}
    \caption{Block diagram of the CN-CBF safety filter. The robot and obstacle states are used to compute the necessary values and derivatives used by the standard CBF-QP safety filter.}
    \label{fig:cn_cbf_safety_filter}
\end{figure}

The final stage of the proposed approach is to integrate the CN-CBF into a CBF-QP safety filter defined in \cref{eq:cbf_qp}. The CN-CBF model provides the CBF value and gradients with respect to the robot and obstacle states. However, the CBF constraint formulation \cref{eq:cbf_constr} also requires $f(\mathrm{x}), g(\mathrm{x})$ and $\{\dot {\mathrm{o}}_i \}_{i=1}^M$, which we can compute in parallel to the CN-CBF block. The time derivative of the $i$th obstacle state at time $t$ is approximated using a backward-differentiation scheme\footnote{If the perception module provides $\dot{\mathrm{o}}_i(t)$ directly, this block can be excluded from the safety filter. Also, more advanced estimation methods can be used in the presence of high noise.}:
\begin{equation} \label{eq:o_dot_approx}
    \dot{\mathrm{o}}_i(t) \approx \frac{\mathrm{o}_i(t) - \mathrm{o}_i(t - \delta t)}{\delta t},
\end{equation}
where $\delta t$ is the time period between the two successive observations.  The overall architecture of the safety filter is presented in \cref{fig:cn_cbf_safety_filter}.

\section{EXPERIMENTAL RESULTS}
\label{subsec:experiments}

We evaluate the proposed method in simulation experiments for two types of mobile robots - a ground robot and a quadrotor. In the following subsection, we provide details on the implementation of the experiments\footnote{Code: \url{https://github.com/bojan-derajic/CN-CBF}}.

\subsection{Simulation Experiments}
\label{subsec:sim_experiments}

The proposed CN-CBF safety filter and the baseline methods are evaluated in the Gazebo simulator. The testing environment consists of pedestrians walking within a rectangular space between random goal positions in a cyclic fashion. The agents are controlled by HuNavSim \cite{perez-higueras_hunavsim_2023}, which implements the social force model (SFM). Besides, the agents do not try to avoid the robot, meaning that collision avoidance must be ensured by the robot. 

The \textbf{ground robot} is the first robot type that we analyze in our experiments. This robot is modeled as a kinematic unicycle with state vector $\mathrm{x} = \left[ x_r, y_r, \theta_r \right]^\top$, where $x_r$ and $y_r$ are positional coordinates and $\theta_r$ is the heading angle. The control vector includes the linear speed $v_r$ and the angular speed $\omega_r$, i.e. $\mathrm{u} = \left[ v_r, \omega_r \right]^\top$, while the dynamics equation is
\begin{equation} \label{eq:kin_unicycle}
    \dot{\mathrm{x}} =
    \begin{bmatrix}
        v_r \cos(\theta_r) & v_r \sin(\theta_r) & \omega_r
    \end{bmatrix}^\top
\end{equation}
with control limits $v_r \in [0.25, 1.0]$ and ${\omega_r \in [-0.8, 0.8]}$.

The motion of a pedestrian (i.e., an obstacle) is described by the unicycle model with a state vector $\mathrm{o} = \left[ x_o, y_o, \theta_o, v_o \right]^\top$ and input vector $\mathrm{d} = \omega_o$, where $\omega_o \in [-0.35, 0.35]$. This choice is motivated by the observation that pedestrians in open spaces most of the time walk toward their goal positions with roughly constant speed and a certain change in orientation necessary to avoid other people. This model can be seen as a more robust constant velocity (CV) model that allows for a change in motion direction. Regarding state and input limits, one can collect a dataset of pedestrian motion in a crowd and compute distributions of speed, turning rate, etc. Then, the state and input limits in the HJ reachability can be chosen to cover the major part of those distributions, achieving a trade-off between robustness and conservativeness. 

We define $\mathrm{z} = \left[ x_{rel}, y_{rel}, \theta_{rel}, v_o \right]^\top$ as a relative state vector, where $x_{rel}$ and $y_{rel}$ are relative position coordinates and $\theta_{rel}$ is the relative angle expressed in the robot's frame of reference\footnote{We could also express it in the world frame, but the relative dynamics $\zeta$ could not be formulated as a function of $\mathrm{z}$ directly in this particular case.}, and $v_o$ is the obstacle's speed. This choice of relative states results in the following relative dynamics:
\begin{equation}
    \dot{\mathrm{z}} =
    \begin{bmatrix}
    -v_r + v_o\cos\theta_{rel} + \omega_r y_{rel} \\
    v_o \sin\theta_{rel} - \omega_r x_{rel} \\
    \omega_o - \omega_r \\
    0
    \end{bmatrix}.
\end{equation}

\begin{figure}
    \centering
    \includegraphics[width=0.75\linewidth]{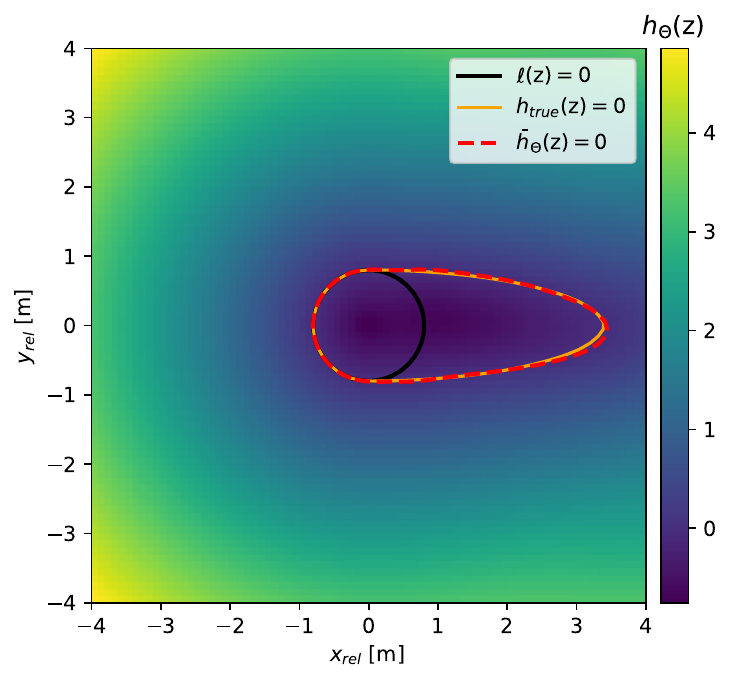}
    \caption{Slice of the learned neural CBF for relative dynamics in the case of the ground robot ($\theta_{rel} = \pi; v_o = 1.2$). The safe region predicted by $\bar{h}_{\Theta}(\mathrm{z})$ never intersects the failure region specified by $\ell(\mathrm{z})$ and closely resembles the safety boundary represented by $h_{\text{true}}(\mathrm{z})$.}
    \label{fig:n_cbf_slice_gr}
\end{figure}

A collision occurs when the relative distance between the robot and obstacle is at most $R_{\min}=R_r+R_o$, where $R_r$ and $R_o$ denote their radii. Accordingly, we define the failure function as $\ell(\mathrm{z})=\sqrt{x_{rel}^2+y_{rel}^2}-R_{\min}$. The residual function $r_{\Theta}(\mathrm{z})$ is represented by an MLP with 4 inputs, 6 hidden layers (48-48-24-24-12-12 units) using sin activations, and one softplus output. Training data are generated with the hj\_reachability\footnote{\url{https://github.com/StanfordASL/hj_reachability}} Python package on an $80\times80\times20\times10$ state-space grid. The model is trained for $10,000$ epochs with Adam. Computing the HJ value function takes $\sim$10 seconds, and training takes $\sim$5.5 minutes on a single GTX 3090 GPU, making the approach substantially more efficient than other learning-based methods. In comparison, \cite{derajic_residual_2025} requires $\sim$20 h for data generation and $\sim$15 h for training in the same setting. The model has 4,837 parameters, versus $1.28\times10^6$ values required by a lookup-table approach. \cref{fig:n_cbf_slice_gr} visualizes a positional slice of the neural CBF for $\theta_{rel}=\pi$ and $v_o=1.2$. Its zero-level set does not intersect the failure region and closely matches the true zero-level set.

We compare our approach against several advanced MPC-based local planners with different representations of safety constraints that have been shown effective in practice: SDF-MPC \cite{zhang_optimization-based_2021}, DCBF-MPC \cite{zeng_safety-critical_2021}, VO-MPC \cite{piccinelli_mpc_2023} and RNTC-MPC \cite{derajic_residual_2025}. As the nominal controller for the proposed CN-CBF safety filter we use an MPC planner without a perception module, meaning that it drives the robot toward the reference state without any safety constraints. All MPC planners and the CBF safety filter are implemented using CasADi \cite{andersson_casadi_2019} and wrapped as ROS 2 nodes. For solving NLP problems in MPC planners we use the IPOPT solver, while for QP problems in the CBF filter we use qpOASES. The MPCs run at a control frequency of 20 Hz and a prediction horizon of 2 s (sampling time 0.1 s), while the safety filter runs at 250 Hz. As the extended class-$\mathcal{K}$ function in \cref{eq:cbf_constr}, we use a linear function $\alpha\left(h(\mathrm{x})\right) = k h(\mathrm{x})$ with $k = 4.0$, while the smoothing parameter in \cref{eq:eta} is set to $\beta = 4.0$. These parameters are simply tuned using a grid search approach.

For the experiments, we first generate 100 random scenarios for 5, 10 and 15 pedestrians, and then test all the methods for the same set of scenarios. The pedestrians are spawned at random locations and move between random goal positions within a rectangular area, thus keeping the effective obstacle density constant over time. The task is to navigate the robot from the start to the goal position without any collisions. As the main metric, we consider the success rate, i.e., the percentage of completed experiments without collision. Additionally, we report path length and time to reach the goal as additional indicators of motion quality.

\begin{figure*}[!tbh]
  \centering

  \begin{minipage}[t]{0.32\textwidth}
    \centering
    \includegraphics[width=\linewidth]{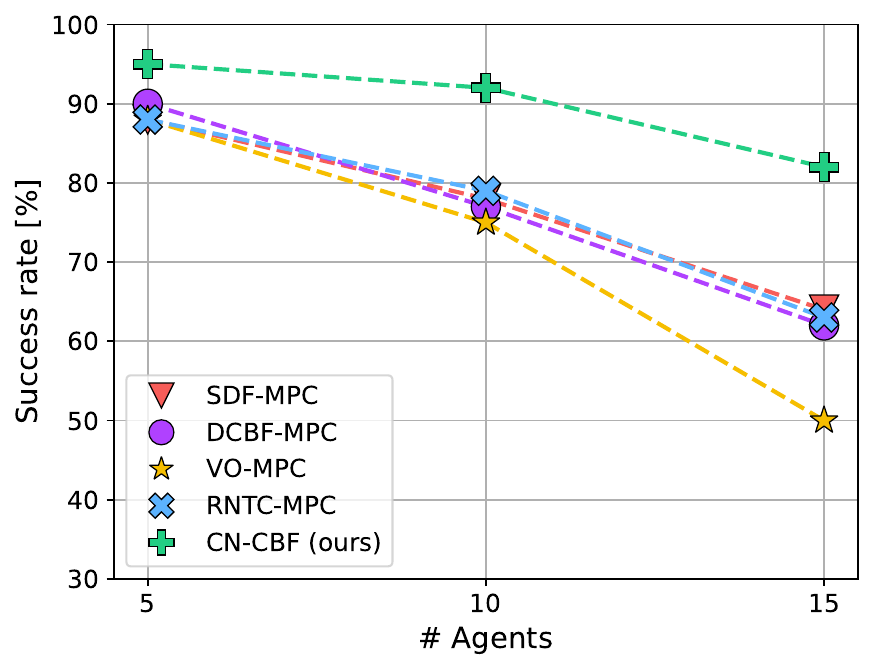}
    
    \scriptsize (a) Success rate
    \label{fig:success_rate_sim_gr}
  \end{minipage}\hfill
  \begin{minipage}[t]{0.32\textwidth}
    \centering
    \includegraphics[width=\linewidth]{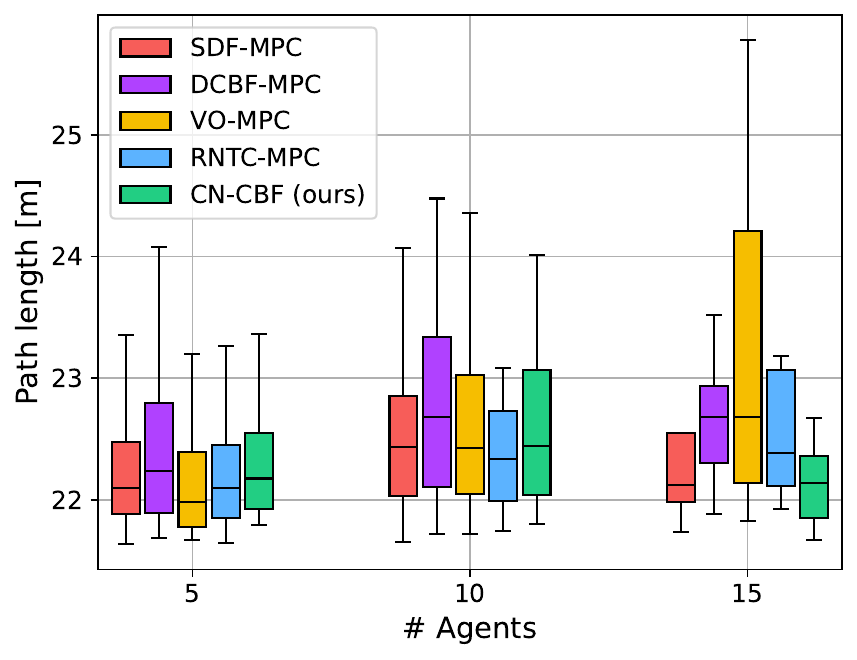}
    
    \scriptsize (b) Path length
    \label{fig:path_length_sim_gr}
  \end{minipage}\hfill
  \begin{minipage}[t]{0.32\textwidth}
    \centering
    \includegraphics[width=\linewidth]{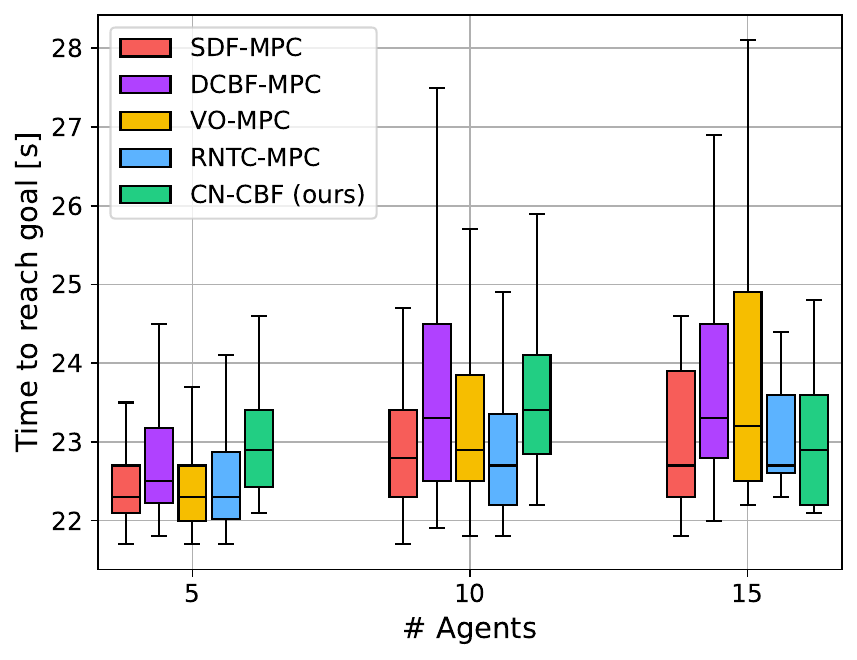}
    
    \scriptsize (c) Time to reach goal
    \label{fig:time_to_reach_goal_sim_gr}
  \end{minipage}

  \caption{Results from the simulation experiments with the ground robot model.}
  \label{fig:sim_results_gr}
\end{figure*}

\begin{figure*}
    \centering
    \includegraphics[width=\linewidth]{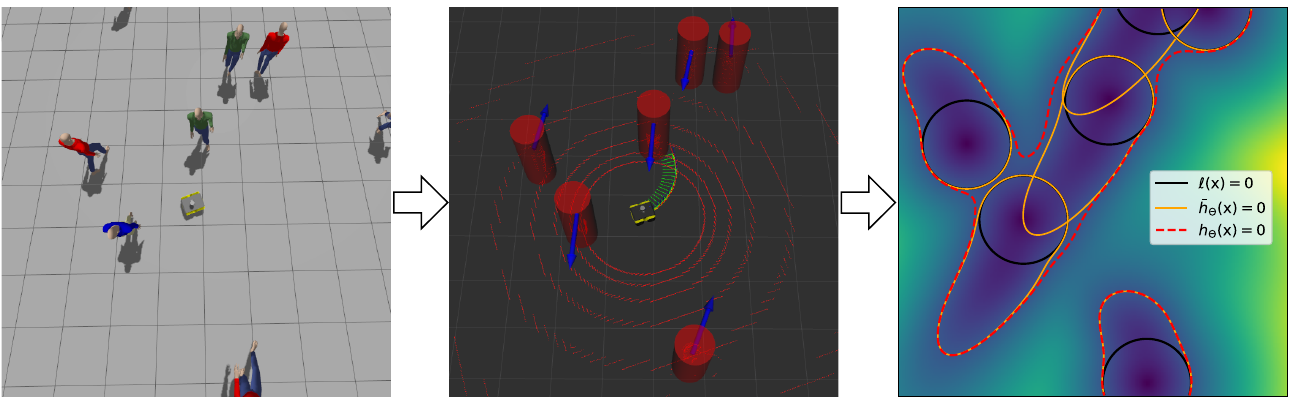}
    \caption{An example scene from the simulation experiments with the ground robot model, which includes the 3D simulator (left), visualization of the local perception (middle), and the corresponding slice of the CN-CBF backprojected to the world frame (right).}
    \label{fig:sample_scene_gr}
\end{figure*}

The results are presented in \cref{fig:sim_results_gr}. The proposed CN-CBF method clearly achieves higher success rates in all cases, with the performance gap widening significantly as obstacle density increases. Moreover, path length and time-to-goal are comparable to, or better than, the baselines. This is important because higher success rates can also result from overly conservative motion. In \cref{fig:sample_scene_gr}, we visualize one sample scenario from the ground robot simulation experiments.

We hypothesize that the success-rate decrease primarily stems from increasing error of the composite CBF relative to the true joint HJ value function as obstacle density grows. Nevertheless, the performance is noticeably better than that of approaches that directly approximate the joint value function, such as RNTC-MPC \cite{derajic_residual_2025}. Other reasons for performance degradation include a limited sensing range and errors in the assumed obstacle dynamics, e.g., when pedestrians abruptly change direction in their motion. 

The \textbf{quadrotor} is the second robot type we analyze in our experiments and is modeled as a 2D double integrator with state vector $\mathrm{x} = \left[ x_r, y_r, v_{x,r}, v_{y,r} \right]^\top$ and control vector $\mathrm{u} = \left[ a_{x,r}, a_{y,r} \right]^\top$. The dynamics of the robot is
\begin{equation} \label{eq:2d_doub_int}
    \dot{\mathrm{x}} =
    \begin{bmatrix}
        v_{x,r} & v_{y,r} & a_{x,r} & a_{y,r}
    \end{bmatrix}^\top
\end{equation}
where $v_{x,r}, v_{y,r} \in [-2, 2]$ and $a_{x,r}, a_{y,r} \in [-2, 2]$.

In this case, the pedestrians are also modeled as 2D double integrators with state vector $\mathrm{o} = \left[ x_o, y_o, v_{x,o}, v_{y,o} \right]^\top$ and input vector $\mathrm{d} = \left[ a_{x,o}, a_{y,o} \right]^\top$ where $v_{x,o}, v_{y,o} \in [-1.5, 1.5]$ and $a_{x,o}, a_{y,o} \in [-1.0, 1.0]$. This obstacle dynamics choice is essentially equivalent to the pedestrian model used in the ground robot case since $v_{x,o} = v_o \cos(\theta_o)$ and $v_{y,o} = v_o \sin(\theta_o)$. However, here we can define the relative state simply as $\mathrm{z} = \mathrm{o} - \mathrm{x}$ with all states being expressed in the world frame, resulting in 2D double integrator relative dynamics:
\begin{equation}
    \dot{\mathrm{z}} =
    \begin{bmatrix}
    v_{x, o} - v_{x, r} \\
    v_{y, o} - v_{y, r} \\
    a_{x, o} - a_{x, r} \\
    a_{y, o} - a_{y, r}
    \end{bmatrix}.
\end{equation}

For the neural model of relative CBF, we use exactly the same MLP architecture and training hyperparameters as in the case of the ground robot. The training data is generated in the same way, but for a state-space grid of shape ${80 \times 80 \times 20 \times 20}$. Also, other implementation details and hyperparameters are the same as in the previous case.

We compare against two CBF baselines suited to this robot dynamics: collision cone CBF (C3BF) \cite{tayal_control_2024} and higher-order CBF (HO-CBF) \cite{xiao_high-order_2022}. The C3BF defines the safety constraint in the velocity space with the aim of preventing the relative velocity vector from entering the collision cone. The HO-CBF uses an SDF function as a CBF and enforces safety by constraining higher-order time derivative (in this case second-order). In both cases, the $\kappa$ functions are linear functions with linear coefficients $k = 3.0$ for C3BF and $k_1 = 3.0, k_2 = 2.0$ for HO-CBF, obtained via a grid-search approach.

\begin{figure*}[!tbh]
  \centering

  \begin{minipage}[t]{0.32\textwidth}
    \centering
    \includegraphics[width=\linewidth]{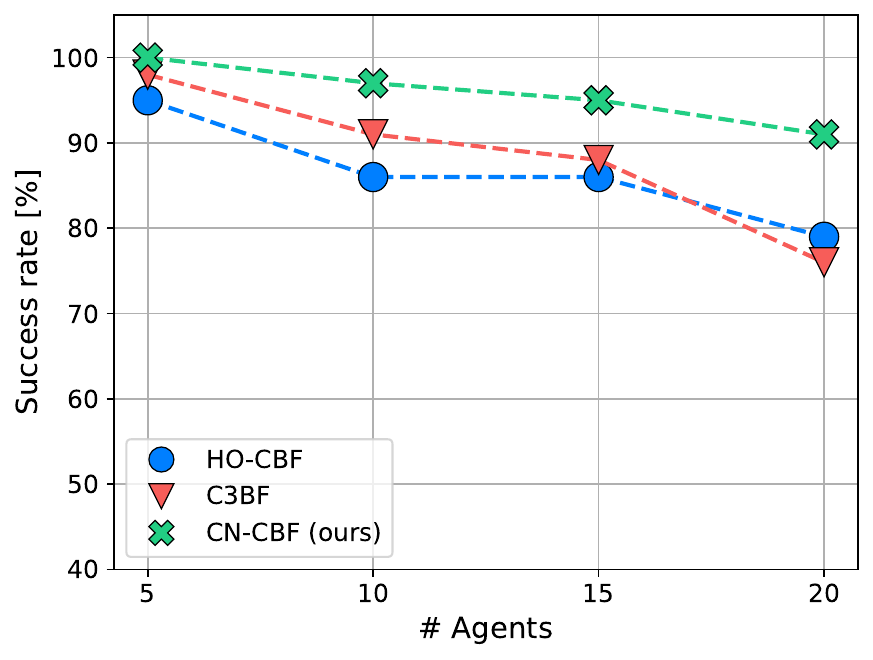}

    \scriptsize (a) Success rate
    \label{fig:success_rate_qq}
  \end{minipage}\hfill
  \begin{minipage}[t]{0.32\textwidth}
    \centering
    \includegraphics[width=\linewidth]{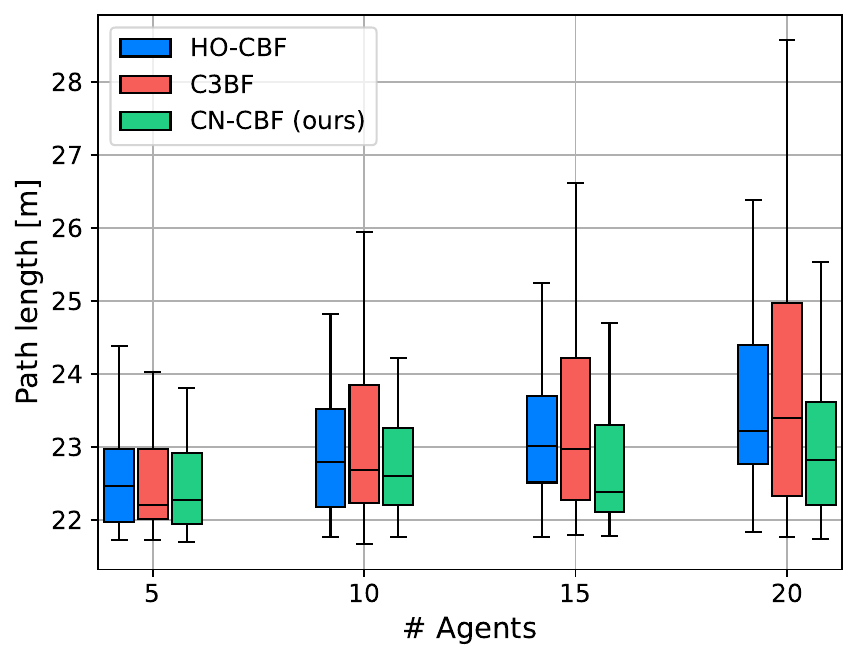}

    \scriptsize (b) Path length
    \label{fig:path_length_sim_qq}
  \end{minipage}\hfill
  \begin{minipage}[t]{0.32\textwidth}
    \centering
    \includegraphics[width=\linewidth]{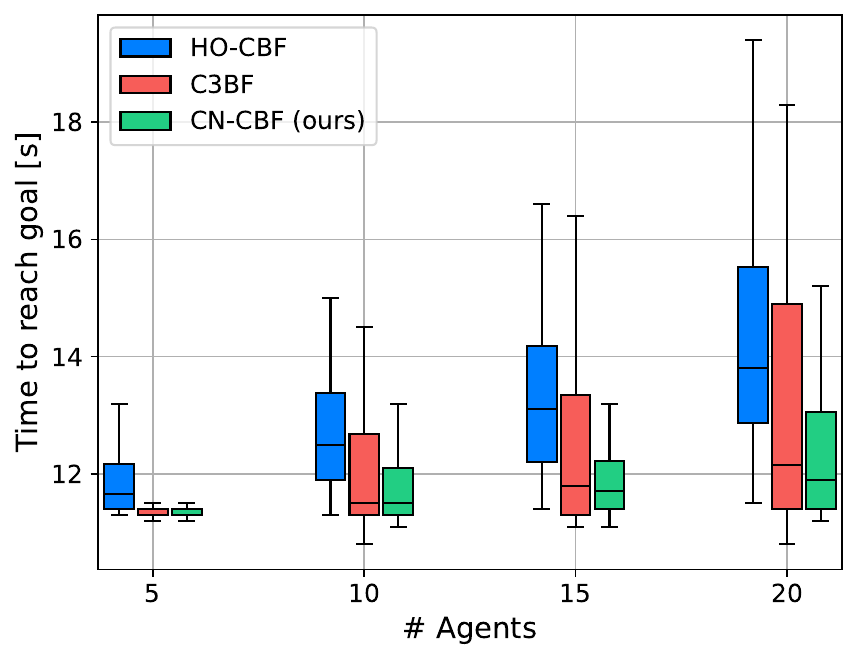}

    \scriptsize (c) Time to reach goal
    \label{fig:time_to_reach_goal_sim_qq}
  \end{minipage}

  \caption{Results from the simulation experiments with the quadrotor model.}
  \label{fig:sim_results_qq}
\end{figure*}

The nominal controller is an LQR, and as in the ground robot case, we generate 100 random scenarios for 5, 10, 15, and 20 pedestrians and test each method on the same set of scenarios. Again, the task is to navigate from the start to the goal position without any collision. The results from these experiments are presented in \cref{fig:sim_results_qq}. Again, it can be seen that our method achieves the best performance, both in terms of success rates and motion efficiency.  

\subsection{Hardware Experiments}
\label{subsec:hardw_experiments}

\begin{figure}
    \centering
    \includegraphics[width=0.9\linewidth]{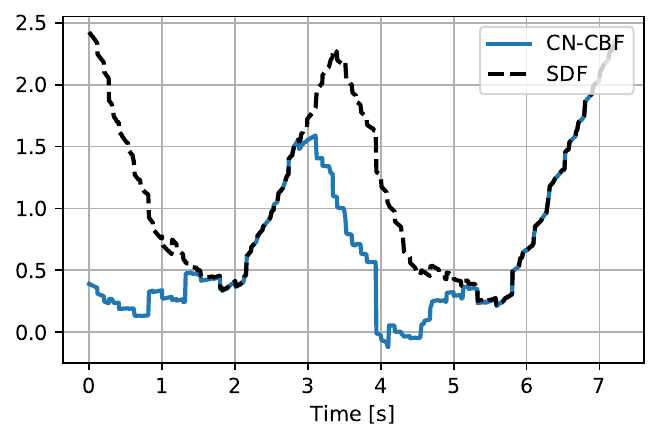}
    \caption{CN-CBF and SDF values over time in the hardware experiments with the ground robot during successful collision avoidance.}
    \label{fig:cn_cbf_gr_hw}
\end{figure}

The proposed CN-CBF method is deployed on a ground robot and a quadrotor in hardware experiments to demonstrate real-time feasibility and practical utility\footnote{For a visual demonstration, see the supplemental video.}. For the ground robot, we use a last-mile delivery robot equipped with 3D LiDAR, a stereo camera, and perception software for detecting moving objects. As in the simulations, the nominal controller is an MPC planner without safety constraints, and collision avoidance relies entirely on the safety filter. The robot and obstacle dynamics are also the same as in the simulations, and all computation is performed onboard. In \cref{fig:cn_cbf_gr_hw}, we show the CN-CBF and SDF values over time during successful pedestrian collision avoidance. The CBF value reaches small negative values due to model mismatch and noisy measurements, but no collision occurs because a small buffer zone is added to the SDF to compensate for these uncertainties.

\begin{figure*}[!tbh]
  \centering

  \begin{minipage}[t]{0.32\textwidth}
    \centering
    \includegraphics[width=\linewidth]{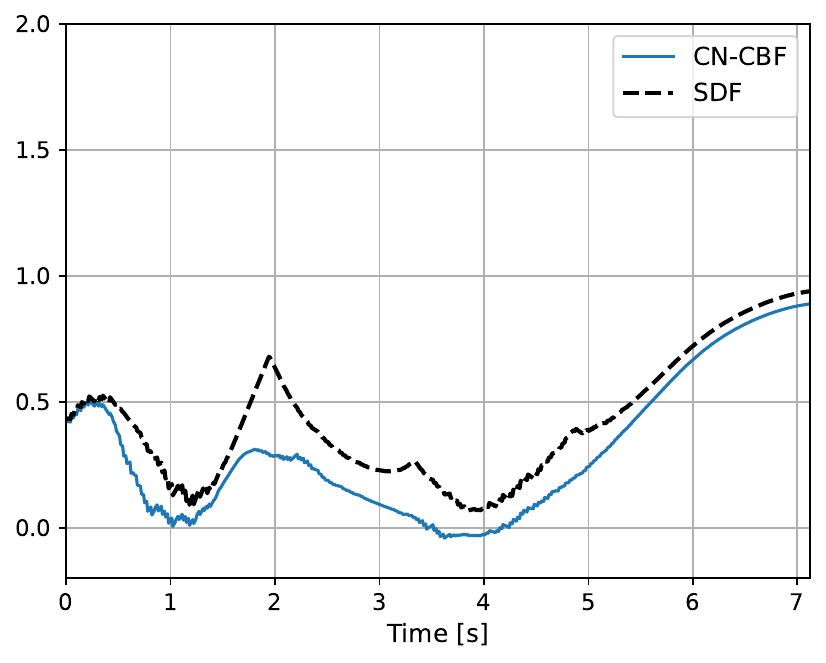}

    \scriptsize (a) CN-CBF (ours)
    \label{fig:cn_cbf_qq_hw}
  \end{minipage}\hfill
  \begin{minipage}[t]{0.32\textwidth}
    \centering
    \includegraphics[width=\linewidth]{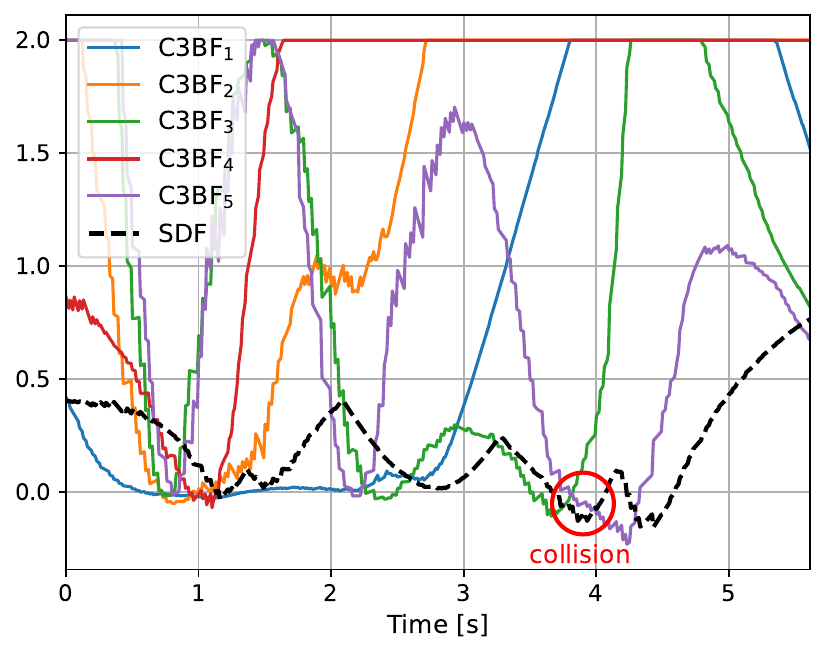}

    \scriptsize (b) C3BF \cite{tayal_control_2024}
    \label{fig:c3bf_qq_hw}
  \end{minipage}\hfill
  \begin{minipage}[t]{0.32\textwidth}
    \centering
    \includegraphics[width=\linewidth]{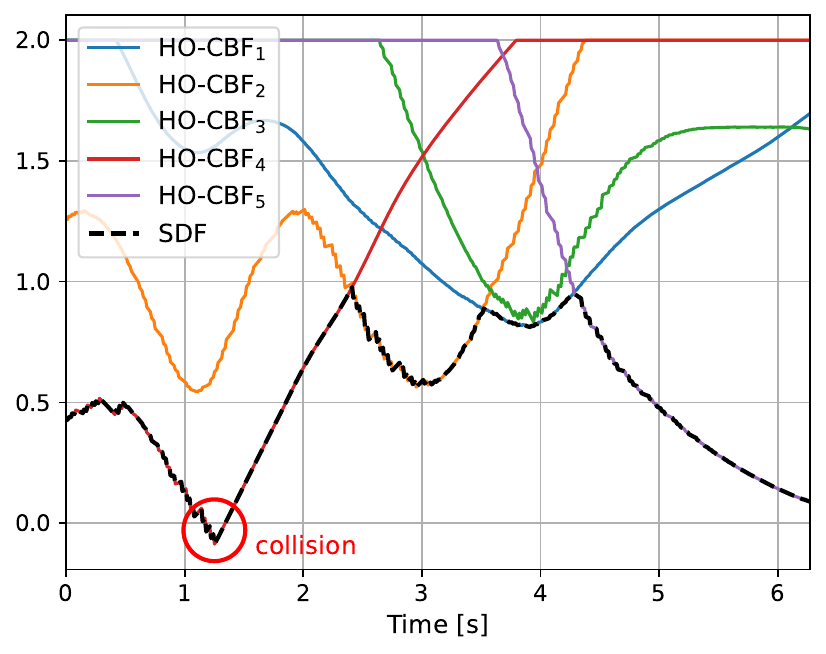}

    \scriptsize (c) HO-CBF \cite{xiao_high-order_2022}
    \label{fig:ho_cbf_qq_hw}
  \end{minipage}

  \caption{CBF and SDF values for different methods obtained in hardware experiments with the quadrotor.}
  \label{fig:hw_results_qq}
\end{figure*}

We also conduct hardware experiments with a Crazyflie quadrotor, comparing CN-CBF method against C3BF and HO-CBF in a collision-avoidance task with five moving obstacles. The robot and obstacle dynamics, and the nominal controller are the same as in the simulations. In this case obstacles are other drones moving between predefined waypoints. The results are visualized in \cref{fig:hw_results_qq}, which shows the time course of the CBF and SDF values\footnote{The data after collision in the case of the baselines correspond to recovering behavior.}. Only our method successfully prevents a collision, while the two baselines collide. Note that the baselines form a CBF per obstacle, while our method constructs only one composite neural CBF.

\section{LIMITATIONS AND FUTURE WORK}
\label{sec:limitations}

In this work, we obtain the true HJ value function using a numerical toolbox, which is limited to relatively low-dimensional systems due to the curse of dimensionality. To improve on this, one can use neural solvers like \cite{bansal_deepreach_2021} or even combine it with MPC \cite{feng_bridging_2025} to obtain an approximate solution for a higher-dimensional relative dynamics. Also, we assume certain obstacle dynamics for the given environment. However, in some cases, it might not be possible to formulate simple obstacle dynamics, and a more flexible approach would be necessary. The residual parameterization excludes modeled failure states from the learned safe set, but does not by itself certify CBF-condition satisfaction or closed-loop safety under model and perception errors. 

For future work, we aim to extend the proposed method to heterogeneous environments with different types of obstacles, where neural CBFs defined for different relative dynamics would be combined into a single composite CBF. In addition, the goal is to improve flexibility and safety by replacing the aggregation function with a trainable model, thereby enabling a better approximation of the true joint CBF.

\section{CONCLUSIONS}

In this paper, we propose a novel CBF design method, called CN-CBF, suited for safe robot navigation in dynamic environments. The method is based on the idea of composite CBFs, where multiple neural CBFs are combined into a single CBF using an aggregation function. The individual CBFs approximate the value function obtained by applying HJ reachability analysis to the relative dynamics between the robot and an obstacle. In this way, the relative CBFs approximately recover the maximal safe set, while the failure function used in HJ reachability remains stationary. Moreover, by using a residual architecture for the individual CBFs, the estimated safe set is guaranteed not to intersect the observed failure set. The method is extensively evaluated in simulation experiments and demonstrated in real-world experiments for two types of robots - a ground robot and a quadrotor. The results show notably improved success rates without increasing motion conservativeness or computational complexity.


\bibliographystyle{IEEEtran}
\bibliography{references}

\begin{thebibliography}{10}
\providecommand{\url}[1]{#1}
\csname url@samestyle\endcsname
\providecommand{\newblock}{\relax}
\providecommand{\bibinfo}[2]{#2}
\providecommand{\BIBentrySTDinterwordspacing}{\spaceskip=0pt\relax}
\providecommand{\BIBentryALTinterwordstretchfactor}{4}
\providecommand{\BIBentryALTinterwordspacing}{\spaceskip=\fontdimen2\font plus
\BIBentryALTinterwordstretchfactor\fontdimen3\font minus \fontdimen4\font\relax}
\providecommand{\BIBforeignlanguage}[2]{{%
\expandafter\ifx\csname l@#1\endcsname\relax
\typeout{** WARNING: IEEEtran.bst: No hyphenation pattern has been}%
\typeout{** loaded for the language `#1'. Using the pattern for}%
\typeout{** the default language instead.}%
\else
\language=\csname l@#1\endcsname
\fi
#2}}
\providecommand{\BIBdecl}{\relax}
\BIBdecl

\bibitem{hsu_safety_2024}
K.-C. Hsu, H.~Hu, and J.~F. Fisac, ``\BIBforeignlanguage{en}{The {Safety} {Filter}: {A} {Unified} {View} of {Safety}-{Critical} {Control} in {Autonomous} {Systems}},'' \emph{\BIBforeignlanguage{en}{Annual Review of Control, Robotics, and Autonomous Systems}}, vol.~7, pp. 47--72, 2024.

\bibitem{nagumo_uber_1942}
M.~Nagumo, ``Über die {Lage} der {Integralkurven} gewöhnlicher {Differentialgleichungen},'' \emph{Proceedings of the Physico-Mathematical Society of Japan. 3rd Series}, vol.~24, pp. 551--559, 1942.

\bibitem{ames_control_2019}
A.~D. Ames, S.~Coogan, M.~Egerstedt, G.~Notomista, K.~Sreenath, and P.~Tabuada, ``Control {Barrier} {Functions}: {Theory} and {Applications},'' in \emph{European {Control} {Conference} ({ECC})}, 2019, pp. 3420--3431.

\bibitem{ames_control_2014}
A.~D. Ames, J.~W. Grizzle, and P.~Tabuada, ``Control barrier function based quadratic programs with application to adaptive cruise control,'' in \emph{{IEEE} {Conference} on {Decision} and {Control}}, 2014, pp. 6271--6278.

\bibitem{lin_one_2026}
A.~Lin, S.~Peng, and S.~Bansal, ``One {Filter} to {Deploy} {Them} {All}: {Robust} {Safety} for {Quadrupedal} {Navigation} in {Unknown} {Environments},'' \emph{IEEE Transactions on Robotics}, vol.~42, pp. 545--560, 2026.

\bibitem{hwang_safe_2024}
S.~Hwang, I.~Jang, D.~Kim, and H.~J. Kim, ``\BIBforeignlanguage{en}{Safe {Motion} {Planning} and {Control} for {Mobile} {Robots}: {A} {Survey}},'' \emph{\BIBforeignlanguage{en}{International Journal of Control, Automation and Systems}}, vol.~22, no.~10, pp. 2955--2969, 2024.

\bibitem{brito_model_2019}
B.~Brito, B.~Floor, L.~Ferranti, and J.~Alonso-Mora, ``Model {Predictive} {Contouring} {Control} for {Collision} {Avoidance} in {Unstructured} {Dynamic} {Environments},'' \emph{IEEE Robotics and Automation Letters}, vol.~4, no.~4, pp. 4459--4466, 2019.

\bibitem{samavi_sicnav_2025}
S.~Samavi, J.~R. Han, F.~Shkurti, and A.~P. Schoellig, ``{SICNav}: {Safe} and {Interactive} {Crowd} {Navigation} {Using} {Model} {Predictive} {Control} and {Bilevel} {Optimization},'' \emph{IEEE Transactions on Robotics}, vol.~41, pp. 801--818, 2025.

\bibitem{derajic_residual_2025}
B.~Derajic, M.-K. Bouzidi, S.~Bernhard, and W.~Hönig, ``\BIBforeignlanguage{en}{Residual {Neural} {Terminal} {Constraint} for {MPC}-based {Collision} {Avoidance} in {Dynamic} {Environments}},'' in \emph{\BIBforeignlanguage{en}{Proceedings of {The} 9th {Conference} on {Robot} {Learning}}}, vol. 305.\hskip 1em plus 0.5em minus 0.4em\relax PMLR, 2025, pp. 1452--1469.

\bibitem{dawson_safe_2023}
C.~Dawson, S.~Gao, and C.~Fan, ``Safe {Control} {With} {Learned} {Certificates}: {A} {Survey} of {Neural} {Lyapunov}, {Barrier}, and {Contraction} {Methods} for {Robotics} and {Control},'' \emph{IEEE Transactions on Robotics}, vol.~39, no.~3, pp. 1749--1767, 2023.

\bibitem{robey_learning_2020}
A.~Robey, H.~Hu, L.~Lindemann, H.~Zhang, D.~V. Dimarogonas, S.~Tu, and N.~Matni, ``Learning {Control} {Barrier} {Functions} from {Expert} {Demonstrations},'' in \emph{{IEEE} {Conference} on {Decision} and {Control}}, 2020, pp. 3717--3724.

\bibitem{liu_safe_2022}
S.~Liu, C.~Liu, and J.~Dolan, ``\BIBforeignlanguage{en}{Safe {Control} {Under} {Input} {Limits} with {Neural} {Control} {Barrier} {Functions}},'' in \emph{\BIBforeignlanguage{en}{Conference on {Robot} {Learning} ({CoRL})}}, 2022.

\bibitem{chen_control_2025}
T.~Chen, A.~Swann, J.~Yu, O.~Shorinwa, R.~Murai, M.~Kennedy, and M.~Schwager, ``A {Control} {Barrier} {Function} for {Safe} {Navigation} with {Online} {Gaussian} {Splatting} {Maps},'' in \emph{{IEEE} {International} {Conference} on {Robotics} and {Automation} ({ICRA})}, 2025, pp. 11\,758--11\,765.

\bibitem{so_how_2024}
O.~So, Z.~Serlin, M.~Mann, J.~Gonzales, K.~Rutledge, N.~Roy, and C.~Fan, ``How to {Train} {Your} {Neural} {Control} {Barrier} {Function}: {Learning} {Safety} {Filters} for {Complex} {Input}-{Constrained} {Systems},'' in \emph{{IEEE} {International} {Conference} on {Robotics} and {Automation} ({ICRA})}, 2024, pp. 11\,532--11\,539.

\bibitem{choi_robust_2021}
J.~J. Choi, D.~Lee, K.~Sreenath, C.~J. Tomlin, and S.~L. Herbert, ``Robust {Control} {Barrier}-{Value} {Functions} for {Safety}-{Critical} {Control},'' in \emph{{IEEE} {Conference} on {Decision} and {Control}}, 2021, pp. 6814--6821.

\bibitem{derajic_orn-cbf_2026}
B.~Derajić, S.~Bernhard, and W.~Hönig, ``{ORN}-{CBF}: {Learning} {Observation}-conditioned {Residual} {Neural} {Control} {Barrier} {Functions} via {Hypernetworks},'' in \emph{{IEEE} {International} {Conference} on {Robotics} and {Automation} ({ICRA})}, 2026.

\bibitem{harms_safe_2025}
M.~Harms, M.~Jacquet, and K.~Alexis, ``Safe {Quadrotor} {Navigation} {Using} {Composite} {Control} {Barrier} {Functions},'' in \emph{{IEEE} {International} {Conference} on {Robotics} and {Automation} ({ICRA})}, 2025, pp. 6343--6349.

\bibitem{safari_time-varying_2024}
A.~Safari and J.~B. Hoagg, ``Time-{Varying} {Soft}-{Maximum} {Control} {Barrier} {Functions} for {Safety} in an {A} {Priori} {Unknown} {Environment},'' in \emph{American {Control} {Conference} ({ACC})}, 2024, pp. 3698--3703.

\bibitem{yu_sequential_2023}
H.~Yu, C.~Hirayama, C.~Yu, S.~Herbert, and S.~Gao, ``Sequential {Neural} {Barriers} for {Scalable} {Dynamic} {Obstacle} {Avoidance},'' in \emph{2023 {IEEE}/{RSJ} {International} {Conference} on {Intelligent} {Robots} and {Systems} ({IROS})}, 2023, pp. 11\,241--11\,248.

\bibitem{lindemann_control_2019}
L.~Lindemann and D.~V. Dimarogonas, ``Control {Barrier} {Functions} for {Signal} {Temporal} {Logic} {Tasks},'' \emph{IEEE Control Systems Letters}, vol.~3, no.~1, pp. 96--101, 2019.

\bibitem{bansal_hamilton-jacobi_2017}
S.~Bansal, M.~Chen, S.~Herbert, and C.~J. Tomlin, ``Hamilton-{Jacobi} reachability: {A} brief overview and recent advances,'' in \emph{{IEEE} {Conference} on {Decision} and {Control}}, 2017, pp. 2242--2253.

\bibitem{molnar_composing_2023}
T.~G. Molnar and A.~D. Ames, ``Composing {Control} {Barrier} {Functions} for {Complex} {Safety} {Specifications},'' \emph{IEEE Control Systems Letters}, vol.~7, pp. 3615--3620, 2023.

\bibitem{perez-higueras_hunavsim_2023}
N.~Pérez-Higueras, R.~Otero, F.~Caballero, and L.~Merino, ``{HuNavSim}: {A} {ROS} 2 {Human} {Navigation} {Simulator} for {Benchmarking} {Human}-{Aware} {Robot} {Navigation},'' \emph{IEEE Robotics and Automation Letters}, vol.~8, no.~11, pp. 7130--7137, 2023.

\bibitem{zhang_optimization-based_2021}
X.~Zhang, A.~Liniger, and F.~Borrelli, ``Optimization-{Based} {Collision} {Avoidance},'' \emph{IEEE Transactions on Control Systems Technology}, vol.~29, no.~3, pp. 972--983, 2021.

\bibitem{zeng_safety-critical_2021}
J.~Zeng, B.~Zhang, and K.~Sreenath, ``Safety-{Critical} {Model} {Predictive} {Control} with {Discrete}-{Time} {Control} {Barrier} {Function},'' in \emph{American {Control} {Conference} ({ACC})}, 2021, pp. 3882--3889.

\bibitem{piccinelli_mpc_2023}
N.~Piccinelli, F.~Vesentini, and R.~Muradore, ``{MPC} {Based} {Motion} {Planning} {For} {Mobile} {Robots} {Using} {Velocity} {Obstacle} {Paradigm},'' in \emph{European {Control} {Conference} ({ECC})}, 2023, pp. 1--6.

\bibitem{andersson_casadi_2019}
J.~A.~E. Andersson, J.~Gillis, G.~Horn, J.~B. Rawlings, and M.~Diehl, ``\BIBforeignlanguage{en}{{CasADi}: a software framework for nonlinear optimization and optimal control},'' \emph{\BIBforeignlanguage{en}{Mathematical Programming Computation}}, vol.~11, no.~1, pp. 1--36, 2019.

\bibitem{tayal_control_2024}
M.~Tayal, R.~Singh, J.~Keshavan, and S.~Kolathaya, ``Control {Barrier} {Functions} in {Dynamic} {UAVs} for {Kinematic} {Obstacle} {Avoidance}: {A} {Collision} {Cone} {Approach},'' in \emph{American {Control} {Conference} ({ACC})}, 2024, pp. 3722--3727, iSSN: 2378-5861.

\bibitem{xiao_high-order_2022}
W.~Xiao and C.~Belta, ``High-{Order} {Control} {Barrier} {Functions},'' \emph{IEEE Transactions on Automatic Control}, vol.~67, no.~7, pp. 3655--3662, 2022.

\bibitem{bansal_deepreach_2021}
S.~Bansal and C.~J. Tomlin, ``{DeepReach}: {A} {Deep} {Learning} {Approach} to {High}-{Dimensional} {Reachability},'' in \emph{{IEEE} {International} {Conference} on {Robotics} and {Automation} ({ICRA})}, 2021, pp. 1817--1824, iSSN: 2577-087X.

\bibitem{feng_bridging_2025}
Z.~Feng, L.~Qiu, and S.~Bansal, ``Bridging {Model} {Predictive} {Control} and {Deep} {Learning} for {Scalable} {Reachability} {Analysis},'' in \emph{Proceedings of {Robotics}: {Science} and {Systems}}, 2025.

\end{thebibliography}

\end{document}